\documentclass[conference,letterpaper]{IEEEtran}

\usepackage{amsmath,amssymb}
\usepackage{graphicx}
\usepackage{float}
\usepackage{microtype}
\usepackage{url}
\usepackage[hidelinks]{hyperref}
\usepackage[backend=biber,style=ieee,sorting=none,maxnames=6,minnames=1]{biblatex}
\title{On the Numerical Reliability of Differentiable Physics-Based Optimization\\for Robotic Material Manipulation}

\author{\IEEEauthorblockN{Xintong Yang\textsuperscript{1}, Minglun Wei\textsuperscript{1}, Yu-Kun Lai\textsuperscript{2}, and Ze Ji\textsuperscript{3,*}}
\IEEEauthorblockA{\textsuperscript{1}School of Engineering, Cardiff University, Cardiff, United Kingdom\\
\textsuperscript{2}School of Computer Science and Informatics, Cardiff University, Cardiff, United Kingdom\\
\textsuperscript{3}College of Mechanical and Electrical Engineering, Hohai University, China\\
\textsuperscript{*}Corresponding author: Ze Ji (z.ji@hhu.edu.cn)}}

\begin{document}
\maketitle

\begin{abstract}
Differentiable physics is increasingly used in robotic material manipulation for system identification, trajectory or skill optimization, demonstration generation, and robot or end-effector design. These applications depend on gradients propagated through long, contact-rich simulation rollouts. We study the numerical reliability of those gradients using two Material Point Method (MPM) system-identification benchmarks derived from elastoplastic and granular manipulation. The benchmarks provide controlled cases for three effects that also arise in broader differentiable physics-based optimization. GPU many-to-one sums whose order depends on thread scheduling changed long-horizon gradients and reversed the sign of one parameter gradient relative to a deterministic reference. Finite-difference checks became less reliable for longer rollouts because repeated-run loss variation grew much faster than the loss change produced by the tested parameter perturbations. Observation and loss definitions changed optimization behaviour and the solution preferred by an independent metric. These results motivate reproducible accumulation, finite-difference validation that compares perturbation-induced loss changes with repeated-run variation, and explicit reporting of objective construction when differentiable simulation is used for robotic optimization.
\end{abstract}

\begin{IEEEkeywords}
differentiable simulation, material point method, physics-based optimization, system identification, deformable object manipulation, reproducibility
\end{IEEEkeywords}

\section{Introduction}
Robotic manipulation of deformable and granular materials appears in food handling, excavation and levelling, laboratory automation, and tool-based interaction. The outcome of these tasks depends on material deformation, friction, flow, and contact with the robot or end effector. Differentiable physics provides gradients through these interactions and has been used to identify material parameters for robotic manipulation \cite{yang2025dpsi,yang2026ddbot}, generate demonstrations for granular-material learning \cite{wei2026physicsinformed}, optimize excavation and levelling skills \cite{wei2025celebi}, and optimize robot shape \cite{ye2026sdrs}. These applications place numerical properties of the simulator directly inside the robotics optimization loop.

\begin{figure}[t]
\centering
\includegraphics[width=0.98\columnwidth]{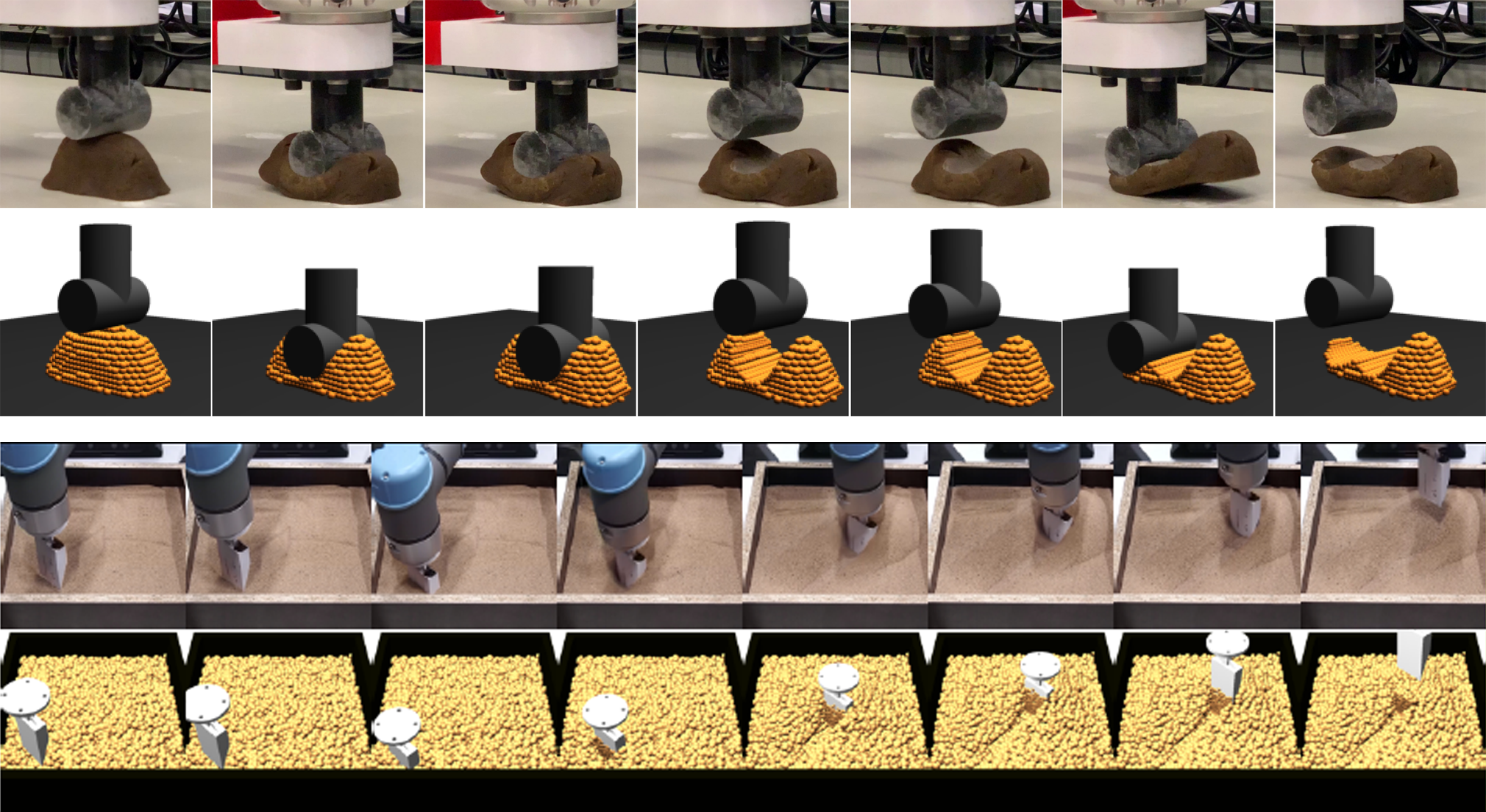}
\caption{Robotic material-manipulation settings motivating this study. Top: Real elastoplastic interaction and the corresponding MPM simulation from the DPSI-derived setting \cite{yang2025dpsi}. Bottom: Real and simulated granular digging from the DDBot-derived setting \cite{yang2026ddbot}.}
\label{fig:robotics}
\end{figure}

System identification, trajectory optimization, skill optimization, and differentiable end-effector design update different variables, while their computational structure is similar. A task loss is evaluated after simulated interaction, and derivatives are propagated backward through the dynamics to the optimized parameters, controls, skills, or geometry. Changes in accumulation order, numerical precision, rollout horizon, or observation discretization can therefore change the local gradient presented to the optimizer. We study these numerical properties as part of the interface between differentiable simulation and robotic optimization.

The Material Point Method (MPM) is useful for robotic interaction with materials undergoing large deformation. Material state is carried by Lagrangian particles and momentum is exchanged through a temporary Eulerian background grid \cite{hu2018mlsmpm}. Each substep performs particle-to-grid (P2G) transfer, grid update and contact handling, followed by grid-to-particle (G2P) transfer. On a GPU, these transfers repeatedly combine many thread-local contributions into one shared grid or gradient entry. We refer to this many-to-one summation as a parallel reduction. Atomic additions allow the contributions to arrive in different thread orders, and floating-point rounding can make the final sum depend slightly on that order \cite{demmel2013reproducible,demmel2015parallel}. Contact and observation losses add further nonlinear and discrete effects. Prior work has also shown that contact formulation and event timing can strongly affect differentiable-simulation gradients \cite{werling2021nimble,zhong2023contactgrad,liu2024softmac}.

Our previous studies provide two controlled robotic testbeds for examining these effects. Differentiable Physics-based System Identification (DPSI) reported multiple parameter solutions and disagreement between Chamfer-distance and Earth-Mover's-distance objectives, with outcomes depending on the loss and initialization \cite{yang2025dpsi}. Differentiable Digging Robot (DDBot) reported exploding and fluctuating gradients and rugged loss landscapes, motivating gradient clipping and line search \cite{yang2026ddbot}. We use DPSI-derived elastoplastic and DDBot-derived granular cases (Fig.~\ref{fig:robotics}) to examine GPU reduction order, finite-difference reliability across rollout length, and the construction of observations and losses. The experiments optimize material parameters. The same numerical operations also occur when gradients are propagated to robot trajectories, skill parameters, or end-effector geometry, which makes the findings relevant to a wider class of differentiable physics-based robotic optimization problems.

\begin{figure*}[t]
\centering
\includegraphics[width=0.99\textwidth]{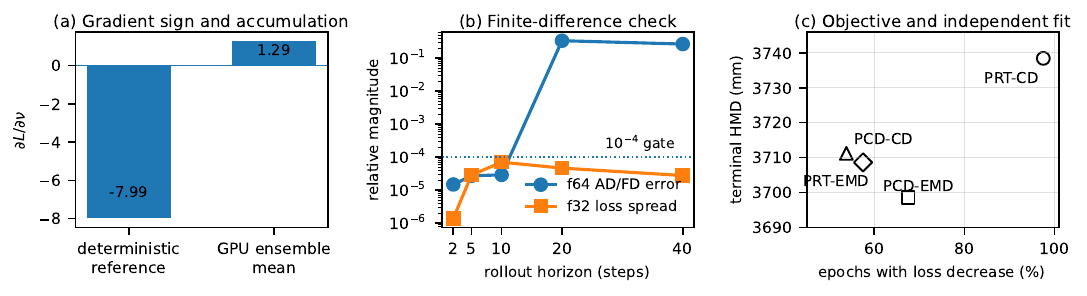}
\caption{Three numerical findings measured in the robotic material-manipulation testbeds. (a) Unordered GPU many-to-one summation changed the sign of a DDBot-derived parameter gradient. (b) Finite-difference agreement in the DPSI-derived case depends on rollout horizon. (c) The objective with the most frequent descent did not give the lowest independent height-map error.}
\label{fig:summary}
\end{figure*}

\section{Method and Study Setup}
\subsection{Benchmark settings}
The DPSI case models elastoplastic material with fixed-corotated elasticity and von Mises plasticity \cite{yang2025dpsi}. The DDBot-derived case models granular material with a Hencky/Saint Venant-Kirchhoff elastic response and Drucker-Prager plasticity \cite{yang2026ddbot}. Both use the same projection-based signed distance field contact formulation. Keeping the contact model fixed allows this study to isolate numerical effects associated with differentiation, reduction order, rollout length, and objective construction. Appendix~\ref{app:mpm} gives the MPM equations, constitutive models, contact formulation, and observation definitions.

We evaluate 32-bit and 64-bit floating point (f32 and f64), several geometric objectives, multiple particle densities for the elastoplastic case, two rollout horizons for the granular case, and deterministic versus ordinary GPU parallel reductions. The complete experiment matrix and optimizer settings are given in Appendix~\ref{app:protocol}. Particle observations (PRT) compare simulated particles directly with a target. Point-cloud observations (PCD) first extract a surface representation. Chamfer distance (CD), Earth Mover's distance (EMD), and height-map distance (HMD) then measure geometric discrepancy. Appendix~\ref{app:mpm} defines these quantities and the point-selection rules.

\subsection{Gradient evaluation and reproducible accumulation}
Gradients are computed with reverse-mode automatic differentiation (AD) through the MPM rollout and compared with central finite differences (FD) over several perturbation sizes. For each perturbation, we compare the resulting change in loss with the spread obtained by repeating an identical forward simulation. This shows whether the FD numerator is large enough to be distinguished from ordinary numerical variation. Long trajectories use checkpointing to limit stored simulation state. Appendix~\ref{app:diff} gives the reverse-mode equations, checkpointing, SVD derivative in detail.

To study parallel reductions, we compare ordinary GPU atomic addition with a deterministic fixed-point alternative. In the ordinary implementation, many threads add particle or gradient contributions to the same destination, so the order of the floating-point additions can change from run to run. In the deterministic implementation, each bounded contribution is mapped to a common integer scale, the integers are added exactly within the allocated range, and the result is converted back to floating point. Appendix~\ref{app:diff} gives the reduction equations, range calculation, implementation details, and a toy numerical example.

\section{Findings}
\subsection{Parallel reductions can change the reverse gradient}
Repeated GPU rollouts first diverge at the P2G parallel reduction, where contributions from many particles are added to the same grid entries. The reverse pass contains two further many-to-one sums: the adjoint of G2P sends contributions back to shared grid entries, and gradients from many particles are combined into shared material-parameter gradients. The differences are initially small, approximately $10^{-7}$ in f32 and $10^{-16}$ in f64, then propagate through later simulation and reverse-mode operations. On the 311-step DDBot-derived granular case, the GPU ensemble mean for $\partial L/\partial\nu$ was $+1.29$, while the deterministic single-threaded CPU reference was $-7.99$ (Fig.~\ref{fig:summary}a). The two signs imply opposite updates of Poisson's ratio.

We replaced the shared reductions involved in particle-grid transfer and parameter-gradient accumulation with deterministic integer reductions, and grouped repeated Chamfer-gradient contributions so they are summed in a fixed order. Five representative configurations were repeated after these changes, including f32 and f64 cases and both benchmark families. All five produced identical state and loss checksums. Appendix~\ref{app:diff} describes the deterministic reductions, and Appendix~\ref{app:results} reports the repeat and accuracy measurements.

\subsection{Finite-difference checks can be unreliable for longer rollouts}
A central finite-difference estimate compares two forward losses evaluated at $\theta+h$ and $\theta-h$. It is a useful external check for AD when the loss difference caused by this parameter perturbation is clearly larger than the variation obtained by repeating the same forward simulation. In the DPSI-derived elastoplastic case, this ratio was 93--806 at two global steps, meaning that the perturbation-induced loss change was much larger than the measured f32 loss spread. At horizons of ten steps or more, the ratio fell to 2--14 because repeated-run loss variation increased much faster than the loss change caused by the tested perturbations.

The f64 finite-difference estimates agreed closely with AD at short horizons: the worst relative discrepancy was at most $1.5\times10^{-5}$ at two global steps and remained below $3\times10^{-5}$ at five and ten steps. At 20 and 40 steps, the finite-difference sequence no longer converged for the worst component under the tested relative perturbation sizes, and the worst AD/FD discrepancy increased to $3.41\times10^{-1}$ and $2.70\times10^{-1}$ (Fig.~\ref{fig:summary}b). Validation cost also increased from 25~s to 178~s. Thus, increasing rollout length did not provide a stronger finite-difference reference in this experiment. Appendix~\ref{app:results} reports the horizon measurements, while Appendix~\ref{app:diff} defines the perturbation-induced loss change and repeated-run loss spread used in this comparison.

\subsection{Observation and loss design change optimization}
DPSI previously reported that CD and EMD can favour different spatial aspects and converge to different parameters \cite{yang2025dpsi}. Under the common numerical setup used here, particle Chamfer distance (PRT-CD) reduced its own objective in 78 of 80 attempted epochs (97.5\%). The corresponding rates were 67.5\% for particle EMD (PRT-EMD), 53.8\% for point-cloud CD (PCD-CD), and 57.5\% for point-cloud EMD (PCD-EMD).

An independent terminal HMD gives a different ranking: mean HMD was 3738.4~mm for PRT-CD and 3698.6~mm for PRT-EMD (Fig.~\ref{fig:summary}c). The particle-density experiment also shows that simulation resolution and observation resolution need not grow together. Increasing the DPSI-derived sampling from 218 to 876 particles increased the number of occupied surface cells from 89 to 151, a factor of 1.70 rather than four. In one 876-particle case, f32 and f64 selected different surface points and changed PCD-EMD by 0.517\%, which is larger than many accepted line-search improvements in the campaign. Appendix~\ref{app:results} reports the complete results.

These measurements show that the loss supplied to the optimizer includes choices made by the observation pipeline as well as the simulated physics. In system identification, this can change the recovered material parameters. In trajectory, policy, or design optimization, it can change the gradient direction associated with controls, contact sequences, or shape variables.

\section{Conclusion}
This study examined numerical reliability in differentiable MPM optimization using elastoplastic and granular system-identification cases as controlled testbeds. The experiments varied floating-point precision, rollout horizon, parallel reduction strategy, particle sampling, and geometric objective while keeping the underlying contact formulation fixed. The appendices provide the MPM formulation, reverse-mode implementation, deterministic accumulation method, complete experimental protocol, and detailed measurements.

There are three findings. First, unordered parallel reductions can propagate small floating-point differences into materially different long-horizon gradients; one DDBot-derived Poisson-ratio gradient changed sign relative to the deterministic reference, while deterministic reductions made all five selected repeat cases bit-identical. Second, finite-difference validation became less reliable as the rollout length increased. In the DPSI-derived horizon experiment, the loss change caused by the tested parameter perturbations was 93-806 times larger than repeated-run f32 variation at two steps, and only 2-14 times larger at horizons of ten steps or more. The f64 finite-difference sequence also stopped converging for the worst component at 20 and 40 steps under the tested perturbation sizes. Third, observation and loss construction changed both optimization behaviour and evaluation ranking; the objective that decreased the most did not achieve the lowest independent HMD, and particle density changed the effective surface observation nonlinearly.

These findings motivate three practices for differentiable physics-based optimization. Many-to-one sums that influence gradients should use a reproducible accumulation strategy when long rollouts can amplify small rounding differences. AD/FD checks should test several perturbation sizes and compare the resulting perturbation-induced loss changes with repeated-forward loss variation; a finite-difference estimate is a weak reference when these quantities are of similar magnitude or when the estimate does not converge as the perturbation size is reduced. The observation, objective, and discrete selection procedures should be treated as part of the optimization specification and reported with enough detail to reproduce their derivatives. These recommendations apply to system identification and to other gradient-based uses of differentiable simulation, including trajectory optimization, contact-rich control, and differentiable robot or end-effector shape optimization, because the same numerical operations connect simulated states to the final optimization variables.

This work is limited to TaiChi-based differentiable MPM implementation and observations collected by two motions (pressing a playdough and scooping a box of sand). Future investigation should be extended to other differentiable programming languages and more diverse contact processes.

\clearpage
\printbibliography[title={References}]
\clearpage
\appendices

\section{MPM and Observation Background}
\label{app:mpm}
\subsection{One MPM substep}
The implementation follows the quadratic Moving Least Squares Material Point Method (MLS-MPM) with an Affine Particle-In-Cell (APIC) velocity representation \cite{hu2018mlsmpm}. Particles carry mass $m_p$, position $x_p$, velocity $v_p$, an affine velocity matrix $C_p$, and deformation gradient $F_p$. Grid nodes are indexed by $i$, have positions $x_i$, and receive particle contributions through interpolation weights $w_{ip}$. A useful simplified view of particle-to-grid transfer is
\begin{align}
 m_i &= \sum_p w_{ip}m_p, \\
 (mv)_i &= \sum_p w_{ip}m_p\left[v_p+C_p(x_i-x_p)\right], \\
 f_i^{\mathrm{int}} &= -\sum_p V_p^0 P_pF_p^T\nabla w_{ip},
 \label{eq:appendix_p2g}
\end{align}
where $V_p^0$ is the reference particle volume and $P_p$ is the first Piola-Kirchhoff stress produced by the constitutive model. The grid velocity is updated using internal force, gravity, boundary conditions, and tool contact,
\begin{equation}
 v_i^{+}=\frac{(mv)_i}{m_i}+\Delta t\left(\frac{f_i^{\mathrm{int}}}{m_i}+g\right),
 \label{eq:appendix_grid}
\end{equation}
for active nodes with nonzero mass. Grid-to-particle transfer interpolates the updated grid velocity back to each particle. The APIC affine matrix is reconstructed from the local grid velocity field, then the deformation and position are advanced approximately as
\begin{equation}
 F_p^{+}=\left(I+\Delta t\,C_p^{+}\right)F_p,
 \qquad
 x_p^{+}=x_p+\Delta t\,v_p^{+}.
 \label{eq:appendix_g2p}
\end{equation}
The actual kernels fuse several algebraic terms for efficiency, while the data flow remains the sequence above. This structure explains why shared accumulation appears in both the forward and backward computations. P2G sends many particle contributions to the same grid node. In reverse mode, differentiating a grid-to-particle gather sends many adjoint contributions back to shared grid entries.

\subsection{Material and contact models}
The DPSI-derived elastoplastic case uses fixed-corotated elasticity with von Mises plasticity, which describes an elastic response followed by irreversible deformation once the deviatoric stress reaches the yield condition \cite{yang2025dpsi}. The DDBot-derived granular case uses a Hencky/Saint Venant-Kirchhoff elastic response with Drucker-Prager plasticity. The Drucker-Prager criterion couples shear resistance to pressure and uses a friction angle to model granular yielding \cite{yang2026ddbot}.

Both benchmark settings use the same projection-based signed distance field (SDF) contact formulation inherited from the earlier simulators. For a query point $x$, the SDF $\phi(x)$ is positive outside the tool and negative inside it. Its spatial gradient gives the outward surface normal $n=\nabla\phi/\|\nabla\phi\|$. When a grid node or particle enters the tool, the relative velocity is decomposed into normal and tangential components. The inward normal component is removed, and the tangential component is modified according to the friction coefficient. Tool motion is included through the relative velocity. Holding this formulation fixed allows the experiments to focus on numerical reliability rather than differences between contact models.

\subsection{Observation operators and losses}
For a simulated particle set $X=\{x_a\}$ and target point set $Y=\{y_b\}$, the symmetric Chamfer distance (CD) used in the benchmark has the form
\begin{equation}
 D_{\mathrm{CD}}(X,Y)=
 \sum_a \min_b\|x_a-y_b\|_2+
 \sum_b \min_a\|y_b-x_a\|_2.
 \label{eq:appendix_cd}
\end{equation}
Earth Mover's distance (EMD) forms a one-to-one pairing between points. We write this pairing as $\pi$, where $\pi(a)=b$ means that simulated point $x_a$ is paired with target point $y_b$:
\begin{equation}
 D_{\mathrm{EMD}}(X,Y)=\min_{\pi}\sum_a\|x_a-y_{\pi(a)}\|_2.
 \label{eq:appendix_emd}
\end{equation}
When the two point sets have the same number of points, $\pi$ is a permutation of the target indices. When their point counts differ, the experiments compare as many pairs as the smaller set contains and record how many target points remain unmatched. This point-count rule matters because changing the number of observed surface points changes both the number of terms in the EMD sum and which target points can participate. Height-map distance (HMD) first projects each point set to a horizontal height grid and sums the per-cell height differences. The DPSI-derived height map covers a $0.11\,\mathrm{m}\times0.11\,\mathrm{m}$ region using $32\times32$ cells, following the previous DPSI evaluation \cite{yang2025dpsi}.

Particle observations (PRT) apply CD or EMD directly to simulated particles. Point-cloud observations (PCD) first apply a surface observer. The observer divides the horizontal region into cells and retains the highest particle in each occupied cell. Its output can change when a cell becomes occupied or empty, or when a different particle becomes the highest one in a cell. Similar discrete choices occur in nearest-neighbour matching when two candidates have equal or nearly equal distances. We record the selected surface winners and point correspondences because changes in these choices can change the loss derivative. These observation details provide context for the stronger variability of PCD objectives reported in Sec.~III-C.

\section{Differentiation and Deterministic Accumulation}
\label{app:diff}
\subsection{Reverse-mode computation in operational form}
Let $s_k$ contain the differentiable simulator state at substep $k$, and let $\theta$ contain the material and contact parameters. One substep and the terminal loss can be written as
\begin{equation}
 s_{k+1}=F_k(s_k,\theta), \qquad L=\ell(s_K).
 \label{eq:appendix_forward}
\end{equation}
Reverse-mode AD propagates the adjoint $\lambda_k=\partial L/\partial s_k$ from the terminal state toward the initial state. With $\lambda_K=\partial\ell/\partial s_K$,
\begin{align}
 \lambda_k &= \left(\frac{\partial F_k}{\partial s_k}\right)^T\lambda_{k+1}, \label{eq:appendix_adjoint}\\
 \nabla_{\theta}L &= \sum_{k=0}^{K-1}\left(\frac{\partial F_k}{\partial \theta}\right)^T\lambda_{k+1}. \label{eq:appendix_paramgrad}
\end{align}
The implementation evaluates these equations with the following sequence for each checkpoint window:
\begin{enumerate}
 \item Recompute the forward states in the current window from its saved boundary state.
 \item Seed the gradient of the geometric loss at the terminal frame of the window.
 \item Execute the gradient kernels in reverse simulation order: advection, G2P, grid update and contact, P2G, and material response.
 \item Apply the hand-written SVD vector-Jacobian product at the material stage because the Taichi SVD operation in the tested build has no generated adjoint.
 \item Copy the adjoint of the first state in the window to the terminal state of the preceding window and continue the reverse computation.
\end{enumerate}
The checkpoint scheme stores $O(W)$ device frames for a window of $W$ global steps and recomputes earlier primal states when needed. The DPSI-derived tape measurement used 1.86~GiB peak device memory, while the DDBot-derived configuration used 2.47~GiB on an 11~GiB RTX 2080 Ti. Grid fields accounted for 99.4\% of the analytic DPSI-derived tape footprint.

For an external gradient check, a scalar parameter component is perturbed by $h$ and evaluated with the central finite difference
\begin{equation}
 g_{\mathrm{FD}}(h)=\frac{L(\theta+h)-L(\theta-h)}{2h}.
 \label{eq:appendix_fd}
\end{equation}
Several values of $h$ are tested. For one parameter $\theta$, we define the perturbation-induced loss change as
\begin{equation}
 \Delta L_{\theta}(h)=\left|L(\theta+h)-L(\theta-h)\right|.
 \label{eq:appendix_fd_signal}
\end{equation}
We also repeat an identical forward evaluation $R$ times and measure its loss spread as
\begin{equation}
 V=\max_{r=1,\ldots,R}L_r-\min_{r=1,\ldots,R}L_r.
 \label{eq:appendix_fd_variation}
\end{equation}
The ratio $\Delta L_{\theta}(h)/V$ is reported only as an interpretation aid. A large value means that the effect of the parameter perturbation is clearly larger than the measured run-to-run variation, while a value near one means that numerical variation can substantially contaminate the finite-difference numerator. In the DPSI-derived sign diagnostic, a factor of ten was used as a conservative case-specific rule before assigning an f32 finite-difference sign; it is not treated as a universal threshold across datasets.

\subsection{Parallel reductions in MPM}
A parallel reduction combines many values produced by different threads into one shared destination. For a grid or gradient entry $j$, the operation has the form
\begin{equation}
 S_j=\sum_{p\in\mathcal{N}(j)} c_{pj},
 \label{eq:appendix_reduction}
\end{equation}
where thread $p$ computes a contribution $c_{pj}$ and all contributions in $\mathcal{N}(j)$ must be added to the same $S_j$. In P2G, $S_j$ can be grid mass or momentum and the contributing threads are particles whose interpolation stencils include node $j$. Reverse mode introduces the same pattern when the adjoint of G2P sends contributions back to shared grid entries and when per-particle derivatives are combined into material-parameter gradients.

On a GPU, these many-to-one sums are commonly implemented with atomic addition. Atomicity prevents two threads from overwriting one another, while the scheduler is still free to choose which contribution is added first. A reduction containing $a$, $b$, and $c$ may therefore be evaluated as $(a+b)+c$ in one execution and $a+(b+c)$ in another. Floating-point addition rounds after each operation and is not exactly associative, so the two orders can differ in their last bits. The first divergence in our isolated MPM test occurred in P2G grid mass and momentum even though all per-particle quantities entering the reduction were bit-identical. The same ordering effect was also present in the reverse G2P scatter and material-parameter reductions.

\subsection{Fixed-point integer accumulation}
The deterministic reduction places all contributions on one common binary scale before adding them. Let $F$ be the number of fractional bits and let $c_i$ be one bounded floating-point contribution. We encode
\begin{equation}
 q_i=\operatorname{round}\!\left(c_i2^F\right),
 \qquad
 Q=\sum_i q_i,
 \qquad
 \widehat{S}=Q2^{-F}.
 \label{eq:appendix_fixedpoint}
\end{equation}
The additions in $Q$ are integer additions, so their result is independent of thread arrival order while the integer range is respected. Quantization occurs once per contribution and conversion back to floating point occurs once after the reduction.

The required integer range can exceed one native field, especially for f64. We therefore represent $Q$ using a high field and a low field,
\begin{equation}
 Q=Q^{\mathrm{hi}}2^W+Q^{\mathrm{lo}},
 \qquad 0\le Q^{\mathrm{lo}}<2^W.
 \label{eq:appendix_highlow}
\end{equation}
This can be read as one wider integer split into two blocks. The implementation uses binary blocks, with carry from the low field transferred to the high field before the value is converted back to floating point.

A small numerical example illustrates the complete reduction. Suppose three threads contribute
\begin{equation}
 c_1=1.25,\qquad c_2=0.875,\qquad c_3=1.50,
\end{equation}
and choose $F=3$, so the common fixed-point scale is $2^F=8$. Equation~\eqref{eq:appendix_fixedpoint} gives
\begin{equation}
 q_1=10,\qquad q_2=7,\qquad q_3=12.
\end{equation}
Any arrival order gives the same integer total $Q=10+7+12=29$, and converting back gives $\widehat S=29/8=3.625$. To show the high/low storage, take the toy field width $W=4$, whose low field stores values from 0 to 15. Then
\begin{equation}
 Q^{\mathrm{hi}}=1,\qquad Q^{\mathrm{lo}}=13,
\end{equation}
because $1\times16+13=29$. If an update makes the low field exceed 15, the overflow is carried into the high field. The real implementation uses much larger binary fields and a scale selected from the reduction bounds; this small example only makes the representation and order independence visible.

The bit budget depends on the maximum number of contributions $n$. With
\begin{equation}
 N=\lceil\log_2 n\rceil,\qquad W=53-N,\qquad B=2W,
 \label{eq:appendix_budget}
\end{equation}
we require $B\ge p+8$, where $p$ is the target floating-point precision (24 significand bits for f32 and 53 for f64). The DDBot-derived P2G reduction with 27,440 particles has $N=15$, $W=38$, and $B=76$, leaving 23 bits beyond f64 precision. The narrowest margin in the complete campaign occurs in the highest-density DPSI-derived PRT-CD reduction, where 31 guard bits remain.

Repeated-source Chamfer gradients need a related change. A generated adjoint may send several pairwise loss terms to the same source particle through floating-point atomic addition. We sort pairs by source particle and store compressed sparse row (CSR) offsets. One source particle then reads and sums its own contiguous block of pair contributions in a fixed order. In the six-launch diagnostic, the generated adjoint produced six distinct gradients, while the CSR gather produced one repeated value.

\section{Experimental Protocol}
\label{app:protocol}
\begin{table}[H]
\caption{Benchmark configurations used in the objective matrix.}
\label{tab:appendix_protocol}
\centering
\small
\setlength{\tabcolsep}{3pt}
\begin{tabular}{p{0.25\columnwidth}p{0.67\columnwidth}}
\hline
Setting & Configuration \\
\hline
DPSI-derived elastoplastic & Fixed-corotated elasticity with von Mises plasticity; 218, 451, or 876 particles; 94 global steps with 50 MPM substeps per step; PRT-CD, PRT-EMD, PCD-CD, and PCD-EMD. \\
DDBot-derived granular & Hencky/Saint Venant-Kirchhoff elasticity with Drucker-Prager plasticity; 27,440 particles; 200 or 311 global steps with 20 MPM substeps per step; HMD and PCD-EMD. \\
\hline
\end{tabular}
\end{table}

The experiment matrix contains 64 primary cells. The DPSI-derived part contains four objectives, two precisions (f32 and f64), three particle densities, one 94-step horizon, and two initialization seeds, giving 48 cells. The DDBot-derived part contains two objectives, two precisions, one particle density, two horizons, and two seeds, giving 16 cells. Five representative cells were repeated for bit-level reproducibility, giving 69 campaign tasks in total. All use the same projection-based SDF contact formulation.

DPSI-derived optimization uses Adam. DDBot-derived optimization uses RMSProp followed by a line-search evaluation of candidate step scales. The DDBot-derived line-search set includes both gradient directions so the experiment can record cases in which the computed direction does not locally reduce the objective. Across the base DDBot-derived cells, 14 of 58 accepted f32 steps and 13 of 42 accepted f64 steps selected the opposite direction.

The particle-density experiment applies only to the DPSI-derived cases. Its sampling densities correspond to 218, 451, and 876 particles. At the initial state, these produce 89, 105, and 151 occupied PCD surface cells. The observed surface representation therefore grows more slowly than the underlying particle count because many added particles project into cells that are already occupied.

\section{Detailed Results}
\label{app:results}
\subsection{Gradient validation across horizon}
\begin{table}[H]
\caption{DPSI-derived horizon experiment. The f64 column reports the worst relative AD/FD discrepancy across the tested material components.}
\label{tab:appendix_horizon}
\centering
\small
\resizebox{\columnwidth}{!}{%
\begin{tabular}{rccc}
\hline
Global steps & f64 worst error & f32 loss spread & FD time \\
\hline
2  & $4.4\times10^{-6}$ to $1.5\times10^{-5}$ & $2.2\times10^{-7}$ to $1.4\times10^{-6}$ & 25~s \\
5  & $2.75\times10^{-5}$ & $2.9\times10^{-5}$ & 38~s \\
10 & $2.93\times10^{-5}$ & $7.1\times10^{-5}$ & 59~s \\
20 & $3.41\times10^{-1}$ & $4.7\times10^{-5}$ & 99~s \\
40 & $2.70\times10^{-1}$ & $2.8\times10^{-5}$ & 178~s \\
\hline
\end{tabular}%
}
\end{table}

Table~\ref{tab:appendix_horizon} gives the measurements summarized in Fig.~\ref{fig:summary}b. At two global steps, $\Delta L_{\theta}(h)/V$ ranged from 93 to 806 across the tested material components. At horizons of ten steps or more, it fell to 2--14 because the repeated-run f32 loss spread increased much faster than the loss change caused by the same relative parameter perturbations. This makes the finite-difference numerator increasingly difficult to distinguish from ordinary numerical variation. At 20 and 40 global steps, the fixed relative perturbation sizes also stopped giving a converged f64 finite-difference sequence for the worst component. The 2-step setting therefore has the lowest runtime, the smallest repeated-forward variation, and the most consistent f64 finite-difference reference among the tested horizons.

\subsection{Objective and particle-density results}
\begin{table}[H]
\caption{Reference-density DPSI-derived optimization behaviour. Descent count is the number of epochs whose post-update objective is lower than the entry value for that epoch, aggregated over two seeds and both precisions. HMD is an independent terminal evaluation.}
\label{tab:appendix_objectives}
\centering
\small
\begin{tabular}{lcc}
\hline
Objective & Descending epochs & Mean HMD (mm) \\
\hline
PRT-CD  & 78/80 (97.5\%) & 3738.4 \\
PRT-EMD & 54/80 (67.5\%) & 3698.6 \\
PCD-CD  & 43/80 (53.8\%) & 3711.2 \\
PCD-EMD & 46/80 (57.5\%) & 3708.6 \\
\hline
\end{tabular}
\end{table}

The full particle-density result is objective-dependent. At f32, PRT-EMD improves from a total loss change of $-0.87\%/-0.48\%$ at the reference density (seed 0/1) to $-2.60\%/-1.26\%$ at $2\times$ density and $-5.21\%/-4.06\%$ at $4\times$ density. PRT-CD changes in the other direction, from $-10.75\%/-2.80\%$ at the reference density to $-0.64\%/-0.30\%$ at $2\times$ and $-0.07\%/-0.60\%$ at $4\times$. These results motivate reporting particle density together with the observation objective.

The PCD surface count grows from 89 to 151 while particle count grows from 218 to 876. Between the first two density levels, 93\% of newly added particles land in an already occupied surface cell; between the second and third levels the figure is 89\%. The particle-density change therefore modifies both the MPM discretization and the observation sampling, while the $32\times32$ PCD grid remains fixed.

\subsection{Runtime and reproducibility measurements}
\begin{table*}[!t]
\caption{Selected computational measurements.}
\label{tab:appendix_runtime}
\centering
\small
\setlength{\tabcolsep}{5pt}
\begin{tabular}{p{0.36\textwidth}p{0.58\textwidth}}
\hline
Case & Runtime or outcome \\
\hline
DPSI 94-step SDF contact, CPU loop vs. GPU kernels & 260.821~s $\rightarrow$ 20.488~s (12.73$\times$) \\
DPSI FD validation, 2 steps vs. 40 steps & 25~s $\rightarrow$ 178~s \\
DDBot HMD epoch, 200 steps & about 115~s; about 154~s with the extended line-search set \\
DDBot PCD-EMD epoch, 311 steps & about 170~s; about 227~s with the extended line-search set \\
DPSI particle-density change & about +1.5\% wall time for denser f32 cases; recorded device memory unchanged \\
Deterministic repeat cells & 5/5 bit-identical across selected f32/f64 and DPSI/DDBot-derived cases \\
\hline
\end{tabular}
\end{table*}

The contact timing isolates a large implementation cost that is separate from the numerical questions in the paper. Both contact backends evaluate the same projection-based SDF contact formulation, while the GPU version removes six host/device frame transfers per MPM substep. The measured speedup at 4,000 particles ranges from 35.1$\times$ on CPU execution of the device-style kernels to 92.9$\times$ on the tested CUDA configuration, depending on precision and device. The 94-step benchmark gives the end-to-end 12.73$\times$ value in Table~\ref{tab:appendix_runtime}.

The deterministic accumulator also improves numerical accuracy against a fixed-order long-double reference in the isolated reduction test. For f32 grid mass, the maximum error was $2.00\times10^{-10}$ with deterministic accumulation and $1.44\times10^{-9}$ for the best of eight atomic runs. The fixed-point representation therefore provides order-independent repetition while retaining adequate precision for the tested reductions.

\subsection{Connection to earlier DPSI and DDBot observations}
The earlier DPSI study observed that CD and EMD may move in opposite directions during optimization and can converge to different parameter values even when the resulting deformations appear similar \cite{yang2025dpsi}. It also reported modelling discrepancies around sharp tool contacts and noted that time integration, step size, and contact handling can contribute to those differences. The objective matrix in this paper adds controlled measurements of observation choice, particle density, precision, and reduction order around the same class of system-identification problem.

The earlier DDBot study reported exploding and fluctuating gradients in long granular trajectories and used gradient clipping and line search during optimization \cite{yang2026ddbot}. The DDBot-derived experiments here isolate one numerical source of such variability by comparing unordered GPU reductions with an order-independent reference. This gives a reproducible benchmark for later studies that change the contact formulation while keeping the remaining numerical controls fixed.

\end{document}